\documentclass{article}
\usepackage{ijcai23}

\usepackage{times}
\usepackage{soul}
\usepackage{url}
\usepackage[hidelinks]{hyperref}
\usepackage[utf8]{inputenc}
\usepackage[small]{caption}
\usepackage{graphicx}
\usepackage{amsmath}
\usepackage{amsthm}
\usepackage{amssymb}
\usepackage{booktabs}
\usepackage{algorithm}
\usepackage{algorithmic}
\usepackage[switch]{lineno}
\usepackage{colortbl}
\usepackage{xcolor}
\usepackage{array}
\usepackage{multirow}
\usepackage{subfig}

\usepackage{bbding}

\title{An Empirical Study on the Language Modal in Visual Question Answering}

\author{
Daowan Peng$^{1,2}$\and
Wei Wei\thanks{Corresponding author.}$^{1,2}$\and
Xian-Ling Mao$^3$\and
Yuanyuan Fu$^{2,4}$\And
Dangyang Chen$^{2,4}$
\affiliations
$^1$Cognitive Computing and Intelligent Information Processing (CCIIP) Laboratory, School of Computer Science and Technology, Huazhong University of Science and Technology\\
$^2$Joint Laboratory of HUST and Pingan Property \& Casualty Research (HPL)\\
$^3$Department of Computer Science and Technology, Beijing Institute of Technology\\
$^4$Ping An Property \& Casualty Insurance Company of China, Ltd\\ 
\emails
\{pengdw, weiw\}@hust.edu.cn,
maoxl@bit.edu.cn,
fuyuanyuan83@gmail.com,
chendangyang273@pingan.com.cn
}

\begin{document}

\maketitle

\begin{abstract}
    Generalization beyond in-domain experience to out-of-distribution data is of paramount significance in the AI domain. Of late, state-of-the-art Visual Question Answering (VQA) models have shown impressive performance on in-domain data, partially due to the language priors bias which, however, hinders the generalization ability in practice. This paper attempts to provide new insights into the influence of language modality on VQA performance from an empirical study perspective. To achieve this, we conducted a series of experiments on six models. The results of these experiments revealed that, 1) apart from prior bias caused by question types, there is a notable influence of postfix-related bias in inducing biases, and 2) training VQA models with word-sequence-related variant questions demonstrated improved performance on the out-of-distribution benchmark, and the LXMERT even achieved a 10-point gain without adopting any debiasing methods. We delved into the underlying reasons behind these experimental results and put forward some simple proposals to reduce the models' dependency on language priors. The experimental results demonstrated the effectiveness of our proposed method in improving performance on the out-of-distribution benchmark, VQA-CPv2. We hope this study can inspire novel insights for future research on designing bias-reduction approaches.
\end{abstract}

\section{Introduction}
\label{sec:intro}
    Visuo-linguistic understanding is an important research topic in the field of multimodal machine learning.
    Vision Language (V+L) tasks, such as image caption \cite{karpathy2015deep}, referring expression comprehension \cite{yu2016modeling}, natural language for visual reasoning \cite{suhr2017corpus}, visual entailment \cite{xie2018visual}, visual commonsense reasoning \cite{zellers2019recognition},  and visual question answering (VQA) \cite{antol2015vqa,anderson2018bottom}, serve as proxy tasks for evaluating the capacity of a multi-modal system to achieve high-level multimodal learning and deeper visuo-linguistic understanding.
    This paper specifically focuses on the VQA task, which has been a long-standing challenge in the domains of computer vision and natural language processing. Previous research has shown that many state-of-the-art VQA models tend to rely excessively on easily learnable language priors instead of effectively reasoning based on the visual content within the images during training \cite{goyal2017making,jing2020overcoming,wen2021debiased}. 
    As a result, these VQA models can achieve decent performance on in-distribution data by capturing superficial correlations in the language modality. However, over-reliance on language priors makes these models fragile and results in poor performance on out-of-distribution (OOD) data in real-world scenarios.

    \begin{figure}[t]
    	\centering
    	\includegraphics[width =.48\textwidth,height=0.19\textwidth]{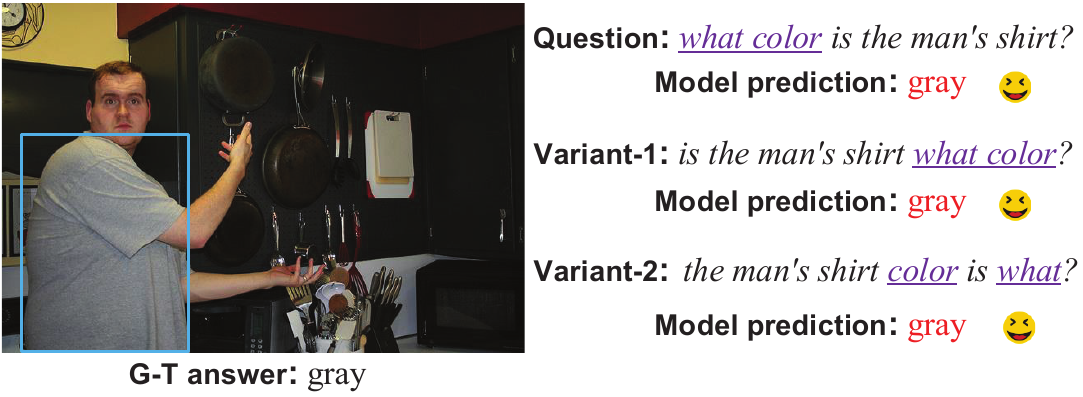}
    	\caption{Here is an example that demonstrates the robustness of VQA models to question disturbances. The purple font is used to indicate the question type. Two kinds of disturbances are shown: Variant-1, which involves exchanging the positions of the prefix (question type) and postfix; and Variant-2, which randomly reorders the words.}  
    	\label{Fig1}
    \end{figure}

    Recently, a broad variety of bias-reduction methods \cite{cadene2019rubi,clark2019dont,liang2021lpf,han2021greedy,yulei2021counterfactual} have been proposed, among which the commonly used approach involves adding a branch to capture language bias. For example, \cite{ramakrishnan2018overcoming} trained a base VQA model along with a question-only adversary to mitigate bias representation by allowing the question-only model to perform poorly. However, some of these methods may introduce extra costs during the inference stage.
    Besides, \cite{yulei2021counterfactual} proposed a novel counterfactual inference framework based on causal effects.
    \cite{han2021greedy} introduced a greedy gradient ensemble de-bias framework, where the bias model is forced to overfit the biased data distribution, allowing the base model to learn the general patterns.
    Apart from the model design side, various methods from the data end have been developed to reduce language priors bias. For instance, HINT \cite{selvaraju2019taking} and SCR \cite{wu2019self-critical} utilize additional annotated data to enhance models' visual-grounding capacity for better performance \footnote{However, it has been revealed that the accuracy improvements of these methods result from the regularization effects \cite{shrestha2020negative}. Besides, collecting such human annotations can be expensive and burdensome.}. 
    CSS \cite{chen:2020counterfactual} generated counterfactual samples by masking the decisive word in the question or crucial object in the image. 
    \cite{liang2020learning} further improved the CSS method by employing contrastive learning to focus on the crucial elements. Thanks to the previous research on debiasing, some progress has been made in addressing the issue of language priors.
    However, in this paper, we aim to provide novel insights regarding the impact of language modality on performance in VQA tasks through empirical investigations.
    In this regard, we conducted a series of confirmatory experimental analysis to investigate prior bias issues. The empirical evidence revealed that, in comparison to the co-occurrence between question types and answers, the co-occurrence between objects and answers could potentially be a more significant factor in contributing to language bias. 
    We also examine the state-of-the-art VQA models' robustness to word-sequence-related disturbance of questions and found that models are resistant to such disturbance to some extent. Figure 1 shows an example.
    Moreover, we found that VQA models trained with variant questions demonstrated higher accuracy in the OOD evaluation. We conducted experiments to analyze the reasons behind this phenomenon and based on these findings, we proposed bias-reduction proposals to alleviate the language bias issue.
    To sum up, the main contributions of this paper are as follows:
    \begin{itemize}
        \item[$\bullet$] 
        We provide empirical evidence demonstrating that language bias in VQA tasks is not solely caused by the co-occurrence of question types and answers, but also by the co-occurrence of visually-grounded concepts and answers, with the latter having a greater impact. Additionally, there may also exist multimodal bias.
        \item[$\bullet$] 
        Extensive experiments reveal that models trained with variant questions outperform those trained with original questions. This improvement is attributed to the disruptions in the word sequence of questions, which impact the model's learning of prior knowledge related to question types, leading to reduced bias-dependency learning.
        \item[$\bullet$] In light of the above findings, we propose de-biasing methods into multiple base VQA models by incorporating variant questions during training. The experimental results demonstrate significant performance enhancements on the VQA-CPv2 benchmark for the base models equipped with our proposed method.
    \end{itemize}

\section{Related Work}
\textbf{Visual Question Answering.} 
As a high-level task that bridges the gap between computer vision and natural language processing, VQA \cite{antol2015vqa,yang2016stacked,agrawal2017vqa,anderson2018bottom,kim2018bilinear,liuy22} has received considerable attention from both the computer vision and natural language processing communities. 
Since the proposal of bottom-up and top-down (UpDn) attention mechanism \cite{anderson2018bottom}, it has been the de-facto standard baseline for the VQA task. \cite{kim2018bilinear} proposed a bilinear attention network (BAN) to efficiently compute multimodal representations. Additionally, \cite{yu2019deep} developed a deep modular co-attention network (MCAN) on top of the powerful Transformer \cite{vaswani2017attention}, which models both intra- and inter-modal interactions simultaneously, making it a powerful baseline for the VQA task.
In addition, pre-trained Vision Language Models (VLMs) \cite{tan2019lxmert,chen2020uniter,su2019vl,zhang2021vinvl,zeng22c,wang2022simvlm} that learn high-level multi-modal representations from large-scale data via a variety of pre-training tasks have demonstrated state-of-the-art performance in many Vision Language (V+L) tasks, including VQA. For instance, the DPT model \cite{liu2022dpt}, which aligns the objectives of the pre-trained visual-language model with the specific requirements of the VQA task, has demonstrated improved generalizability and performance.

\textbf{Bias and Robustness in VQA.} 
The study of robustness in VQA is an important topic, particularly the issue of language bias, which significantly affects the OOD performance in VQA task. As such, an increasing number of bias-mitigation approaches \cite{cadene2019rubi,guo2021adavqa,chen:2020counterfactual,han2021greedy} and benchmarks \cite{agrawal2017vqa,agrawal2018dont,kervadec2021roses} have been proposed. 
\cite{cadene2019rubi} built a question-only branch to capture the unwanted regularities by dynamically adjusting the loss.
\cite{clark2019dont} trained a naive model that relied solely on dataset biases and then used an ensemble approach to incorporate a robust model that focused on other generalized patterns.
\cite{liang2021lpf} added a question-only branch to measure the intensity of language priors and then reshaped the objective function based on the loss of the question-only branch. 
\cite{lao2021language} proposed the LP-Focal loss, which endows the cross-entropy loss with sample-level loss re-weights by building a question-only branch to capture language priors.
\cite{yang2021learning} proposed a CCB method by building content and context branches to focus on local content and global context, respectively. On top of these two branches, a joint loss function with language bias optimizes the prediction.
\cite{kervadec2021roses} suggested that the standard evaluation metric is misleading by the overall accuracy under the unbalanced concepts and questions, thus they proposed a new benchmark consisting of a dataset and a new evaluation metric.  
Apart from the language bias issue, 
\cite{gokhale2020vqa} found that VQA models could answer single questions but struggled to answer logical compositions of multiple such questions. Therefore, they constructed an augmentation of the VQA dataset by collecting logical composition questions, including negation, conjunction, disjunction, and antonyms.
\cite{shah2019cycle} proposed a training scheme by exploiting cycle consistency to regularize the training process, which allows VQA models to become robust to linguistic variations. 
Besides, \cite{kervadec2021transferable} argued that noise and uncertainties in visual inputs are the main bottlenecks in VQA, which prevent the successful learning of reasoning capacities.
SwapMix \cite{gupta2022swapmix} investigated the robustness of VQA models from the perspective of visual context. They swapped some irrelevant objects in the image and found VQA models are not robust for such visual context perturbation, indicating models over-rely on them to make predictions.

\section{Empirical Analysis}
\begin{table*}[!ht]
  \centering
  \renewcommand{\arraystretch}{1.1}
  \setlength{\abovecaptionskip}{0.3cm}
    \begin{tabular}{c|ccccccccccc}
    \hline
    \multirow{2}{*}{Model} & \multicolumn{5}{c}{VQA-CPv2}          &       & \multicolumn{5}{c}{VQAv2} \\
\cline{2-6}\cline{8-12}          & \cellcolor[rgb]{ .929,  .929,  .929}\textit{ques} & \cellcolor[rgb]{ 1,  .949,  .8}\textit{pre-train} & \cellcolor[rgb]{ 1,  .949,  .8}\textit{post-train} & \cellcolor[rgb]{ .886,  .937,  .855}\textit{pre-test} & \cellcolor[rgb]{ .886,  .937,  .855}\textit{post-test} &       & \cellcolor[rgb]{ .929,  .929,  .929}\textit{ques} & \cellcolor[rgb]{ 1,  .949,  .8}\textit{pre-train} & \cellcolor[rgb]{ 1,  .949,  .8}\textit{post-train} & \cellcolor[rgb]{ .886,  .937,  .855}\textit{pre-test} & \cellcolor[rgb]{ .886,  .937,  .855}\textit{post-test} \\
\cline{1-6}\cline{8-12}    Q-only & \cellcolor[rgb]{ .906,  .902,  .902}21.37 & \cellcolor[rgb]{ 1,  .949,  .8}17.34 & \cellcolor[rgb]{ 1,  .949,  .8}\textbf{24.87} & \cellcolor[rgb]{ .886,  .937,  .855}15.22 & \cellcolor[rgb]{ .886,  .937,  .855}\textbf{16.85} &       & \cellcolor[rgb]{ .906,  .902,  .902}45.09 & \cellcolor[rgb]{ 1,  .949,  .8}33.33 & \cellcolor[rgb]{ 1,  .949,  .8}\textbf{35.68} & \cellcolor[rgb]{ .886,  .937,  .855}\textbf{31.73} & \cellcolor[rgb]{ .886,  .937,  .855}25.9 \\
    SAN   & \cellcolor[rgb]{ .906,  .902,  .902}40.7 & \cellcolor[rgb]{ 1,  .949,  .8}22.75 & \cellcolor[rgb]{ 1,  .949,  .8}\textbf{40.35} & \cellcolor[rgb]{ .886,  .937,  .855}20.07 & \cellcolor[rgb]{ .886,  .937,  .855}\textbf{27.16} &       & \cellcolor[rgb]{ .906,  .902,  .902}62.78 & \cellcolor[rgb]{ 1,  .949,  .8}44.93 & \cellcolor[rgb]{ 1,  .949,  .8}\textbf{50.71} & \cellcolor[rgb]{ .886,  .937,  .855}\textbf{39.9} & \cellcolor[rgb]{ .886,  .937,  .855}36.77 \\
    UpDn  & \cellcolor[rgb]{ .906,  .902,  .902}41.53 & \cellcolor[rgb]{ 1,  .949,  .8}26.12 & \cellcolor[rgb]{ 1,  .949,  .8}\textbf{42.3} & \cellcolor[rgb]{ .886,  .937,  .855}21.7 & \cellcolor[rgb]{ .886,  .937,  .855}\textbf{28.75} &       & \cellcolor[rgb]{ .906,  .902,  .902}65.56 & \cellcolor[rgb]{ 1,  .949,  .8}45.19 & \cellcolor[rgb]{ 1,  .949,  .8}\textbf{52.65} & \cellcolor[rgb]{ .886,  .937,  .855}\textbf{40.78} & \cellcolor[rgb]{ .886,  .937,  .855}37.32 \\
    BAN   & \cellcolor[rgb]{ .906,  .902,  .902}41.73 & \cellcolor[rgb]{ 1,  .949,  .8}26.6 & \cellcolor[rgb]{ 1,  .949,  .8}\textbf{28.18} & \cellcolor[rgb]{ .886,  .937,  .855}22.18 & \cellcolor[rgb]{ .886,  .937,  .855}\textbf{37.76} &       & \cellcolor[rgb]{ .906,  .902,  .902}67.07 & \cellcolor[rgb]{ 1,  .949,  .8}37.28 & \cellcolor[rgb]{ 1,  .949,  .8}\textbf{37.87} & \cellcolor[rgb]{ .886,  .937,  .855}39.17 & \cellcolor[rgb]{ .886,  .937,  .855}\textbf{41.19} \\
    LXMERT & \cellcolor[rgb]{ .906,  .902,  .902}40.96 & \cellcolor[rgb]{ 1,  .949,  .8}28.44 & \cellcolor[rgb]{ 1,  .949,  .8}\textbf{43.61} & \cellcolor[rgb]{ .886,  .937,  .855}21.05 & \cellcolor[rgb]{ .886,  .937,  .855}\textbf{36.77} &       & \cellcolor[rgb]{ .906,  .902,  .902}64.51 & \cellcolor[rgb]{ 1,  .949,  .8}44.91 & \cellcolor[rgb]{ 1,  .949,  .8}\textbf{54.37} & \cellcolor[rgb]{ .886,  .937,  .855}40.24 & \cellcolor[rgb]{ .886,  .937,  .855}\textbf{42.23} \\
    MCAN  & \cellcolor[rgb]{ .906,  .902,  .902}43.73 & \cellcolor[rgb]{ 1,  .949,  .8}\textbf{26.31} & \cellcolor[rgb]{ 1,  .949,  .8}26.11 & \cellcolor[rgb]{ .886,  .937,  .855}20.5 & \cellcolor[rgb]{ .886,  .937,  .855}\textbf{34.41} &       & \cellcolor[rgb]{ .906,  .902,  .902}68.65 & \cellcolor[rgb]{ 1,  .949,  .8}\textbf{41.46} & \cellcolor[rgb]{ 1,  .949,  .8}40.56 & \cellcolor[rgb]{ .886,  .937,  .855}39.17 & \cellcolor[rgb]{ .886,  .937,  .855}\textbf{39.87} \\
    \hline
    \end{tabular}%
    \caption{The performance in terms of accuracy (Acc. \%) on the VQA-CPv2 test split and VQAv2 validation split. The \textit{ques} columns indicate models trained and tested with original question; \textit{pre-train} and \textit{post-train} denote models trained on prefix and postfix respectively and tested on the original question; \textit{pre-tes}t and \textit{post-test} refer to models trained on the original question and tested on the prefix and postfix respectively.
    The better results are bold. Note that the LXMERT was fine-tuned for 10 epochs.}
  \label{tab1}%
\end{table*}

\subsection{Task Definition}
    The VQA task has been cast as a classification problem, where given an image, $I$ and a question $Q$, the objective is to predict an answer $\hat{a}$ from all the candidate answers $A$. This prediction is based on the image content and the context of the question.
    Without loss of generality, a VQA model can be formulated as a function transformation $F:(Q,I)\mapsto A$. 
    The objective function $p(.)$ is formulated as:
    \begin{equation}
    \hat{a}=\mathop{\arg\max}_{a\in A}p(a|Q,I;\Theta),
    \end{equation}
    where $\Theta$ denotes the model parameters. 
    The common solution to predict the answer is via the cross-entropy loss
    \begin{equation}
    \begin{aligned}
    \mathcal{L}_{ce} = -\frac{1}{N} \sum_{i}^{N}a_{i}log(p_i)\\
    p_i = Softmax(W  h_i+b),\\
    \end{aligned}
    \end{equation}  
    where $N$ denotes the number of samples, $W$ and $b$ are the learnable matrix and bias, $h_i$ is the fused multi-modal feature.

\subsection{Revisiting Question in VQA}

\subsubsection{Which Contributes More Bias?}
    The language priors bias in the VQA task is generally attributed to the co-occurrence of certain question-types and answers \cite{agrawal2018dont}, and most bias-reduction methods are designed based on this hypothesis.
    In this section, we attempt to verify that the language bias issue is not solely due to the co-occurrence of question-types and answers through empirical analysis.
    To begin, we decompose the question into two parts: the question type (also known as the prefix) and the concepts (which include objects and other visually-grounded words or phrases in the question, also known as the postfix). We then examine their respective contributions to the final accuracy of the model. 
    Intuitively, it is difficult to answer a question correctly if the question is incomplete (\textit{i.e.}, only the prefix or postfix is given). If an incomplete question is answered correctly, it suggests that there is some co-occurrence correlation between the incomplete portion and the corresponding answer, which indicates the presence of bias.
    These experiment settings are as follows.
    \textbf{Dataset}: We selected the widely-used VQAv2 benchmark \cite{goyal2017making} and its OOD benchmark, VQA-CPv2 \cite{agrawal2018dont}. 
    \textbf{Base VQA models}: In the experiment, we chose the most frequently used base models in the VQA task, which include attention-based models such as SAN \cite{yang2016stacked} and UpDn \cite{anderson2018bottom}), bilinear attention network, BAN \footnote{We use the 4-layers version of BAN in this paper.} \cite{kim2018bilinear}, co-attention based model, MCAN \cite{yu2019deep}), multi-modal pre-trained model, LXMERT \cite{tan2019lxmert} and a question-only model (henceforth, Q-only).
    Among them, the LXMERT model uses BERT \cite{devlin2018bert} as the question encoder, while the other models use LSTM \cite{hochreiter1997long} or GRU \cite{cho2014learning} as question encoders.
    \textbf{Validation mode}: We conducted two types of verification. The first type involved training models with original questions and evaluating them on either the prefix or postfix. The second type involved training models with either prefix or postfix and evaluating them on the original questions. All text inputs were padded or truncated to a fixed length using a predetermined character. Moreover, for the sake of simplicity in implementation, we used the questions with the prefix removed as postfix in our experiments.

    The experimental results are presented in Table 1, and several important findings can be derived from these results. 
    The results for the first type of verification mode are displayed in columns highlighted with a light green background in Table 1. We observed that all models with postfix inputs achieved better performance than those with prefix inputs on the VQA-CPv2 test split, particularly for the BAN, LXMERT, and MCAN models, where the postfix inputs contributed significantly more than the prefix. On the VQAv2 dataset, the BAN, LXMERT, and MCAN models performed slightly better with postfix inputs, while the prefix inputs resulted in better performance for the remaining models. The experimental results for the second type of verification mode are shown in columns against a light yellow background in Table 1. As can be seen, when models were trained with the postfix, their accuracy performances were better than those trained with the prefix for all the models except MCAN, which showed slightly lower performance on both VQA-CPv2 and VQAv2. The results from both verification modes in Table 1 indicate that postfix contribute more to the bias issue than prefixes.
    Moreover, we noticed that certain models (\textit{e.g.,} Q-only, UpDn and LXMERT) trained with postfix even outperformed those trained with the complete question on the VQA-CPv2 dataset. We conjecture that this could be attributed to the differences in question-type distributions between the training and test splits. Consequently, some models that do not learn information about the prefix of questions during the training process may exhibit better performance on VQA-CPv2.

\subsubsection{Are There Any Other Kinds of Bias?}
    Also, Table 1 demonstrates that all models perform better than the Q-only model. Apart from the difference in model design, the Q-only model only takes the question as input, whereas others incorporate both the question and image as inputs. This allows these models to potentially depend on the co-occurrence between visual objects in the image and the keywords in the question. Therefore, we speculate that the bias issue exists not only in the language modality but also across multimodalities due to the fact that incomplete questions cannot be answered.

\subsubsection{The Robustness to Variant Question}
    Robustness has always been a crucial concern in machine learning. \cite{cui2022pert} proposed a novel pre-training paradigm for language models. Their approach involves predicting the original order of perturbed words in text, aiming to enhance the model's resilience to the text modality and improve its ability to comprehend text semantics.
    Inspired by their work, this paper investigated the robustness of state-of-the-art VQA models to disruptions in the questions. To achieve this, a series of confirmatory experiments were conducted to evaluate the robustness of VQA models.
    Specifically, we conducted the experiments on three types of variant questions. Given a question such as ``\textit{what color is the flower?}" with the prefix  ``\textit{what color is}", we defined three kinds of variant questions as follows:
    \begin{itemize}
    \item[$\bullet$] variant-1: = {\tt Concate}(postfix, prefix), \textit{i.e.,} exchange the positions of the prefix and postfix, resulting in a variant such as ``\textit{the flower what color is?}''
    
    \item[$\bullet$] variant-2: = {\tt Random}(question), \textit{i.e.,} shuffle the order of the words in the question randomly, resulting one of the possibilities as ``\textit{the flower color is what?}''
    
    \item[$\bullet$] variant-3: = {\tt Inverse}(question), \textit{i.e.,} inverse the word sequence of the question, making the variant as ``\textit{flower the is color what?}''
    \end{itemize}
    To evaluate the model's robustness to the variant questions, we define an evaluation metric $Rob$ as follow,
    \begin{equation} %linewidth
        \%Rob =\frac{N_{rv,rq}}{N_{rq}} \times 100\%,
    \end{equation}
    where $N_{rq}$ represents the number of correct predictions for the original questions, and $N_{rv,rq}$ the number of both original questions and their variants that are correctly answered. 
    
    \begin{table}[t]
    	\centering
    	\setlength\tabcolsep{1mm}  %修改列间距
      \renewcommand{\arraystretch}{1.02}
    	\begin{tabular}{c|c|ccrcc}
    		\hline
    		\multirow{2}{*}{Model} & \multirow{2}{*}{tested with} & \multicolumn{2}{c}{VQA-CPv2} &       & \multicolumn{2}{c}{VQAv2} \\
    		\cline{3-4}\cline{6-7}          &       & Acc.  & $Rob$ &       & Acc.  & $Rob$ \\
    		\hline
    		\multicolumn{1}{c|}{\multirow{4}[2]{*}{Q-only}} & question & \textbf{21.37} & --      &       & \textbf{45.09} &--  \\
    		& variant-1  & 18.30  & 59.5 &       & 33.26 & 61.7 \\
    		& variant-2  & 19.10  & 53.6 &       & 32.73 & 61.0 \\
    		& variant-3  & 15.22 & 40.7 &       & 27.75 & 51.0 \\
    		\cline{1-4}\cline{6-7}    \multirow{4}[2]{*}{SAN} & question & \textbf{40.7} & --      &       & \textbf{62.78} &--  \\
    		& variant-1  & 30.42 & 61.6 &       & 50.05 & 73.7 \\
    		& variant-2  & 28.52 & 53.9 &       & 47.59 & 70.0 \\
    		& variant-3  & 27.43 & 51.4  &       & 44.87 & 65.2 \\
    		\cline{1-4}\cline{6-7}    \multirow{4}[2]{*}{UpDn} & question & \textbf{41.53} & --      &       & \textbf{65.56} &--  \\
    		& variant-1  & 36.86 & 69.1 &       & 55.71 & 79.7 \\
    		& variant-2  & 31.39 & 55.3 &       & 49.83 & 70.8 \\
    		& variant-3  & 28.38 & 47.7 &       & 47    & 66.4 \\
    		\cline{1-4}\cline{6-7}    \multicolumn{1}{c|}{\multirow{4}[2]{*}{BAN}} & question & 41.73 &--       &       & \textbf{67.07} &--  \\
    		& variant-1  & \textbf{42.29} & 79.9 &       & 59.52 & 84.6 \\
    		& variant-2  & 39.87 & 70.1  &       & 55.26 & 78.3 \\
    		& variant-3  & 35.09 & 60.6 &       & 48.42 & 68.1 \\
    		\cline{1-4}\cline{6-7}    \multicolumn{1}{c|}{\multirow{4}[2]{*}{LXMERT}} & question & 40.96 &--       &       & \textbf{64.51} &--  \\
    		& variant-1  & \textbf{43.06} & 87.4 &       & 60.76 & 90.4 \\
    		& variant-2  & 40.17 & 74.7  &       & 56.28 & 82.6 \\
    		& variant-3  & 39.12 & 69.9 &       & 53.96 & 78.6 \\
    		\cline{1-4}\cline{6-7}    \multicolumn{1}{c|}{\multirow{4}[2]{*}{MCAN}} & question & 43.73 &--       &       & \textbf{68.65} &--  \\
    		& variant-1  & \textbf{44.13} & 85.2 &       & 64.8  & 91.7 \\
    		& variant-2  & 42.73 & 74.7 &       & 59.17 & 82.6 \\
    		& variant-3  & 41.68 & 68.9  &       & 57.88 & 80.4 \\
    		\hline
    	\end{tabular}%
    	\caption{The accuracy (Acc.\%) and $Rob$ in terms of different variant models on VQA-CPv2 test split and VQAv2 validation split.}
    	\label{tab2}%
    \end{table}%
    For the selection of VQA models, we have used the ones chosen in the previous section. The experimental results are presented in Table 2. The results on the VQAv2 validation split demonstrate that all models experience varying degrees of performance degradation when evaluated on variant questions. Among them, MCAN and LXMERT show comparable performance on this in-distribution dataset. Furthermore, the results regarding robustness, $Rob$ indicate that all models exhibit the best robustness for variant-1, followed by variant-2, and the worst for variant-3.
    \begin{table*}[ht!]
      \centering
      \setlength{\tabcolsep}{7.5pt}
      \renewcommand{\arraystretch}{1.02} 
        \begin{tabular}{c|c|lcccrcccc}
        \hline
        \multirow{2}{*}{Model} & \multirow{2}{*}{trained with} & \multicolumn{4}{c}{VQA-CPv2}  &       & \multicolumn{4}{c}{VQAv2} \\
    \cline{3-6}\cline{8-11}          &       & All   & Yes/No   & Num   & Other &       & All   & Yes/No   & Num   & Other \\
        \hline
        \multicolumn{1}{c|}{\multirow{4}[1]{*}{Q-only}} & question & \cellcolor[rgb]{ .906,  .902,  .902}21.37 & \cellcolor[rgb]{ .906,  .902,  .902}41.01 & \cellcolor[rgb]{ .906,  .902,  .902}12.14 & \cellcolor[rgb]{ .906,  .902,  .902}13.61 & \cellcolor[rgb]{ .906,  .902,  .902} & \cellcolor[rgb]{ .906,  .902,  .902}{\textbf{45.09}} & \cellcolor[rgb]{ .906,  .902,  .902}69.57 & \cellcolor[rgb]{ .906,  .902,  .902}32.37 & \cellcolor[rgb]{ .906,  .902,  .902}29.81 \\
              & variant-1 & {\textbf{27.80}}\scriptsize\textcolor[rgb]{ 0,  .69,  .314}{+6.43} & 53.64 & 39.32 & 11.09 &       & 41.81 & 68.21 & 30.21 & 24.78 \\
              & variant-2 & 22.41\scriptsize\textcolor[rgb]{ 0,  .69,  .314}{+1.04} & 42.47 & 12.4  & 14.65 &       & 43.9  & 68.55 & 31.99 & 28.29 \\
              & variant-3 & 25.90\scriptsize\textcolor[rgb]{ 0,  .69,  .314}{+4.53}  & 66.02 & 1.34  & 11.62 &       & 32.38 & 65.69 & 1.09  & 15.32 \\
    
        \hline
        \multirow{4}[0]{*}{SAN} & question & \cellcolor[rgb]{ .906,  .902,  .902}40.70 & \cellcolor[rgb]{ .906,  .902,  .902}41.62 & \cellcolor[rgb]{ .906,  .902,  .902}13.14 & \cellcolor[rgb]{ .906,  .902,  .902}47.77 & \cellcolor[rgb]{ .906,  .902,  .902} & \cellcolor[rgb]{ .906,  .902,  .902}{\textbf{62.78}} & \cellcolor[rgb]{ .906,  .902,  .902}78.69 & \cellcolor[rgb]{ .906,  .902,  .902}41.52 & \cellcolor[rgb]{ .906,  .902,  .902}56.31 \\
              & variant-1 & 40.95\scriptsize\textcolor[rgb]{ 0,  .69,  .314}{+0.25} & 56.03 & 15.5  & 40.02 &       & 57.25 & 75.69 & 35.62 & 48.96 \\
              & variant-2 & {\textbf{41.32}}\scriptsize\textcolor[rgb]{ 0,  .69,  .314}{+0.62} & 43.21 & 13.17 & 48.06 &       & 61.47 & 76.96 & 40.76 & 55.17 \\
              & variant-3 & 31.41\scriptsize\textcolor[rgb]{ 0,  .69,  .314}{-9.29} & 40.18 & 12.46  & 32.01 &       & 48.45 & 68.09 & 24.21  & 39.94 \\
        \hline
        \multirow{4}[1]{*}{UpDn} & question & \cellcolor[rgb]{ .906,  .902,  .902}41.53 & \cellcolor[rgb]{ .906,  .902,  .902}42.91 & \cellcolor[rgb]{ .906,  .902,  .902}13.56 & \cellcolor[rgb]{ .906,  .902,  .902}48.55 & \cellcolor[rgb]{ .906,  .902,  .902} & \cellcolor[rgb]{ .906,  .902,  .902}{\textbf{65.56}} & \cellcolor[rgb]{ .906,  .902,  .902}82.87 & \cellcolor[rgb]{ .906,  .902,  .902}44.9 & \cellcolor[rgb]{ .906,  .902,  .902}57.87 \\
              & variant-1 & {\textbf{44.83}}\scriptsize\textcolor[rgb]{ 0,  .69,  .314}{+3.30} & 60.45 & 20.84 & 43.23 &       & 60.37 & 79.86 & 35.21 & 52.21 \\
              & variant-2 & 42.33\scriptsize\textcolor[rgb]{ 0,  .69,  .314}{+0.80} & 44.89 & 13.35 & 48.94 &       & 64.22 & 81.5  & 43.75 & 56.5 \\
              & variant-3 & 33.89\scriptsize\textcolor[rgb]{ 0,  .69,  .314}{-7.64} & 41.22 & 24.9 & 32.52 &       & 50.35 & 71.52 & 29.68 & 39.72\\
              
        \hline
        \multicolumn{1}{c|}{\multirow{4}[0]{*}{BAN}} & question & \cellcolor[rgb]{ .906,  .902,  .902}41.73 & \cellcolor[rgb]{ .906,  .902,  .902}42.72 & \cellcolor[rgb]{ .906,  .902,  .902}13.51 & \cellcolor[rgb]{ .906,  .902,  .902}48.95 & \cellcolor[rgb]{ .906,  .902,  .902} & \cellcolor[rgb]{ .906,  .902,  .902}{\textbf{67.07}} & \cellcolor[rgb]{ .906,  .902,  .902}84.11 & \cellcolor[rgb]{ .906,  .902,  .902}48.2 & \cellcolor[rgb]{ .906,  .902,  .902}59.11 \\

              & variant-1 & 47.19\scriptsize\textcolor[rgb]{ 0,  .69,  .314}{+5.46} & 61.21 & 18.41 & 47.74 &       & 63.81 & 81.39 & 44.67 & 55.51 \\
              & variant-2 & {\textbf{49.92}}\scriptsize\textcolor[rgb]{ 0,  .69,  .314}{+8.19} & 67.92 & 20.72 & 48.49 &       & 63.04 & 81.1  & 42.91 & 54.62 \\
              & variant-3 & 45.87\scriptsize\textcolor[rgb]{ 0,  .69,  .314}{+4.14} & 66.81 & 15.04 & 43.37 &       & 55.19 & 75.51 & 33.01 & 45.61 \\

        \hline
        \multicolumn{1}{c|}{\multirow{4}[2]{*}{LXMERT}} & question & \cellcolor[rgb]{ .906,  .902,  .902}43.29 & \cellcolor[rgb]{ .906,  .902,  .902}46.37 & \cellcolor[rgb]{ .906,  .902,  .902}15.38 & \cellcolor[rgb]{ .906,  .902,  .902}49.34 & \cellcolor[rgb]{ .906,  .902,  .902} & \cellcolor[rgb]{ .906,  .902,  .902}{\textbf{65.67}} & \cellcolor[rgb]{ .906,  .902,  .902}83.31 & \cellcolor[rgb]{ .906,  .902,  .902}46.69 & \cellcolor[rgb]{ .906,  .902,  .902}57.29 \\
 
              & variant-1 & {\textbf{53.66}}\scriptsize\textcolor[rgb]{ 0,  .69,  .314}{+10.37} & 75.21 & 21.4  & 51.22 &       & 65.34 & 83.14 & 45.82 & 56.96 \\
              & variant-2 & 43.57\scriptsize\textcolor[rgb]{ 0,  .69,  .314}{+0.28} & 46.49 & 15.71 & 49.68 &       & 65.29 & 83.19 & 46.35 & 56.7 \\
              & variant-3 & 45.33\scriptsize\textcolor[rgb]{ 0,  .69,  .314}{+2.04} & 51.93 & 22.2  & 48.21 &       & 59.82 & 76.78 & 43.18 & 51.33 \\
    
        \hline
        \multicolumn{1}{c|}{\multirow{4}[2]{*}{MCAN}} & question & \cellcolor[rgb]{ .906,  .902,  .902}43.73 & \cellcolor[rgb]{ .906,  .902,  .902}42.6 & \cellcolor[rgb]{ .906,  .902,  .902}15.69 & \cellcolor[rgb]{ .906,  .902,  .902}52.02 & \cellcolor[rgb]{ .906,  .902,  .902} & \cellcolor[rgb]{ .906,  .902,  .902}{\textbf{68.65}} & \cellcolor[rgb]{ .906,  .902,  .902}85.91 & \cellcolor[rgb]{ .906,  .902,  .902}51.05 & \cellcolor[rgb]{ .906,  .902,  .902}60.17 \\

              & variant-1 & 48.57\scriptsize\textcolor[rgb]{ 0,  .69,  .314}{+4.84} & 52.53 & 26.15 & 52.64 &       & 66.45 & 81.4  & 50.73 & 59.23 \\
              & variant-2 & {\textbf{48.79}}\scriptsize\textcolor[rgb]{ 0,  .69,  .314}{+5.06} & 55.79 & 22.00    & 52.48 &       & 66.65 & 81.93 & 50.61 & 59.27 \\
              &variant-3 & 48.47\scriptsize\textcolor[rgb]{ 0,  .69,  .314}{+4.74} & 65.75 & 18.19 & 47.73 &       & 58.97 & 79.85 & 38.37 & 48.54 \\
        \hline
        \end{tabular}%
        \caption{The performance (in \%) in terms of different variant models on VQA-CPv2 test split and VQAv2 validation split.}
      \label{tab3}%
    \end{table*}%
    \begin{table*}[!ht]
      \centering
      \setlength\tabcolsep{1.2mm} 
      \renewcommand{\arraystretch}{1.05}
        \begin{tabular}{c|ccc|ccc|ccc|ccc|ccc|ccc}
     \hline
        \multirow{2}[4]{*}{} & \multicolumn{3}{c|}{Q-only } & \multicolumn{3}{c|}{SAN } & \multicolumn{3}{c|}{UpDn} & \multicolumn{3}{c|}{BAN} & \multicolumn{3}{c|}{MCAN} & \multicolumn{3}{c}{LXMERT} \\
    \cline{2-19}          & var1  & var2  & var3  & var1  & var2  & var3  & var1  & var2  & var3  & var1  & var2  & var3  & var1  & var2  & var3  & var1  & var2  & var3 \\
        \hline
        \Checkmark $\rightarrow$ \XSolidBrush & 28.5  & \textbf{11.2} & 42.1  & 23.0  & \textbf{10.5} & 44.3  & 20.0  & \textbf{11.3} & 44.6  & 15.5 & \textbf{14.7}  & 23.4  & 14.1  & \textbf{14.1} & 21.6  & \textbf{8.5} & 9.0   & 22.6 \\
        \XSolidBrush $\rightarrow$ \Checkmark & 17.5  & 6.0   & \textbf{17.8} & \textbf{19.1} & 11.7  & 16.6  & \textbf{23.1} & 13.1  & 21.2  & 23.7  & \textbf{28.0} & 26.3  & 24.1  & 24.2  & \textbf{28.9} & \textbf{27.6} & 10.6 & 23.4 \\
        \hline
        \end{tabular}% compared to original models
        \caption{The ratio (in \%) of prediction changes of variant models on VQA-CPv2 test split. The mark \Checkmark $\rightarrow$ \XSolidBrush (lower is better) measures the ratio (based on the predictions of original models) of questions that the original model can answer correctly while the variant models cannot; \XSolidBrush $\rightarrow$ \Checkmark (higher is better)
        represents the opposite case. 
        The var(x) is abbreviated of variant-(x).} 
      \label{tab4}%
    \end{table*}%
    Meanwhile, from the results of the VQA-CPv2 test split, an intriguing observation was that when tested with variant-1 questions, the test accuracy was even higher than that of the original questions in terms of the BAN, MCAN, and LXMERT models. The above experimental results lead us to consider a hypothesis, that is, variant questions may help alleviate language priors dependency.
    To further verify this hypothesis, we conducted another series of experiments. In this experimental setup, we trained VQA models using variant questions as text inputs, and the resulting models are referred to as variant models.
    During the inference stage, we evaluated the performance of the variant models on original questions.
    The experimental results are presented in Table 3. As shown, some variant models achieved comparable performance to that of the original models on VQAv2 dataset. 
    Furthermore, almost all the variant models achieve better performance on VQA-CPv2, except for SAN and UpDn, whose performances degrade when trained with variant-3. 
    The LXMERT, fine-tuned for 20 epochs, even achieved a 10-point gain without adopting any debiasing methods when trained with variant-1 questions.
    Besides, a notable phenomenon can be observed in Table 3, which is that the accuracy changes of different variant models mainly occur in the ``Yes/No" metric, although the changes demonstrated by different models vary.
    
\subsection{Why Did the Performance Improve?}
    To figure out the reasons behind the accuracy improvements from these variant models on VQA-CPv2, we conducted further experimental analysis. Firstly, we performed a fine-grained analysis of the experimental results to investigate how the predictions of the variant models differed from those of the original models. To be more specific, we aimed to determine the number of predictions that changed from correct to incorrect and vice versa for the variant models.
    
    The results are presented in Table 4, from which it can be seen that almost all of the variant-2 models have the smallest proportion of samples that were predicted correctly by the original model but predicted incorrectly by the variant models.
    The corresponding variant-1 model of the LXMERT model performs the best, with the smallest proportion of correct predictions flipped to incorrect ones, and it can also flip 27.6\% of the incorrect predictions to correct ones. Besides, the BAN and MCAN models also demonstrate good performance regarding converting incorrect predictions to correct ones.
    Additionally, we conducted further analysis to examine the question types that correspond to the change in performance of the variant models. The results are depicted in Figure 2.
    It can be observed that the accuracy improvement mainly stems from the ``Yes/No" answer type, and the most frequent question type is \textit{``is there''}. On the other hand, the decreased performance mainly comes from the ``Other" answer type, and the most frequent corresponding question type is \textit{``what"}.
    Based on the previous results, we aimed to investigate further which words in the question the variant models focus on and how this differs from the attention of the original models. 
    To achieve this, one of the most straightforward ways is to compare the feature representations produced by the models and their corresponding variant models for the same question. We visualized the attention weights of the questions with respect to the models and their variant models.
    Specifically, the question was first fed into the model to obtain its feature representation. Then, we mapped the attention to each word of the question. Figure 3 presents two toy examples.
    The subplots (a) and (c) in Figure 3 show that the model trained using the original question input mode places a higher weight on question type, such as the examples \textit{``is this"} and \textit{``what color is"}. 
    The abundance of questions that start with the phrase \textit{``is this"} in the training dataset makes it easier for the models to learn these simple patterns.
    In contrast, for the models trained with variant questions, the aforementioned scenario does not occur, resulting in slightly smoother visualization feature representations, as shown in subplots (b) and (d) of Figure 3.

    \begin{figure}[!t]
    \centering
    	\subfloat[]{\includegraphics[width = 0.235\textwidth]{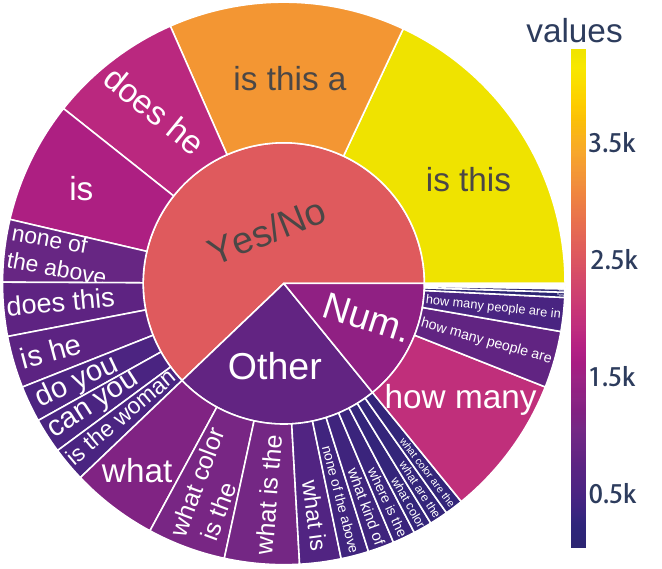}} 
        \label{fig:short-1}
    	\subfloat[]{\includegraphics[width = 0.235\textwidth]{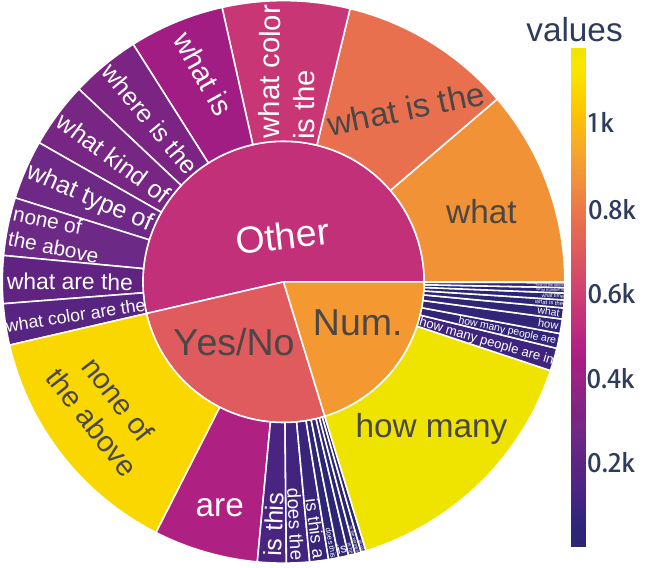}}
        \label{fig:short-2}
        \subfloat[]{\includegraphics[width = 0.235\textwidth]{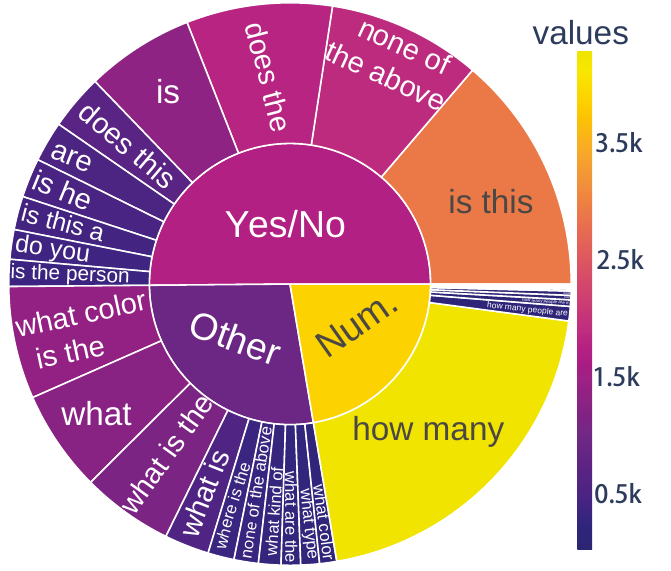}}  
        \label{fig:short-3}
    	\subfloat[]{\includegraphics[width = 0.235\textwidth]{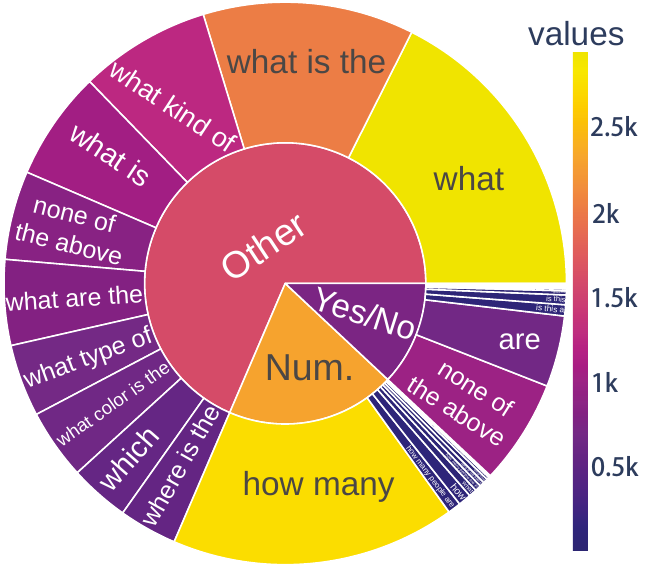}}
        \label{fig:short-4}
    \caption{Statistics on the top-10 question types for each answer type corresponding to prediction-flip samples. The subplots (a) and (b) are statistics with respect to best-performing variant model of LXMERT, while (c) and (d) are with respect to best-performing variant model of UpDn. The first column represents the distribution of question types where the model changed its incorrect predictions to correct predictions, while the second column represents the opposite case.}
    \label{Fig:sunburst}
    \vspace{-0.15cm} 
    \end{figure}
    \begin{figure*}[!ht]
    \centering
    	\subfloat[]{\includegraphics[width = 0.25\textwidth]{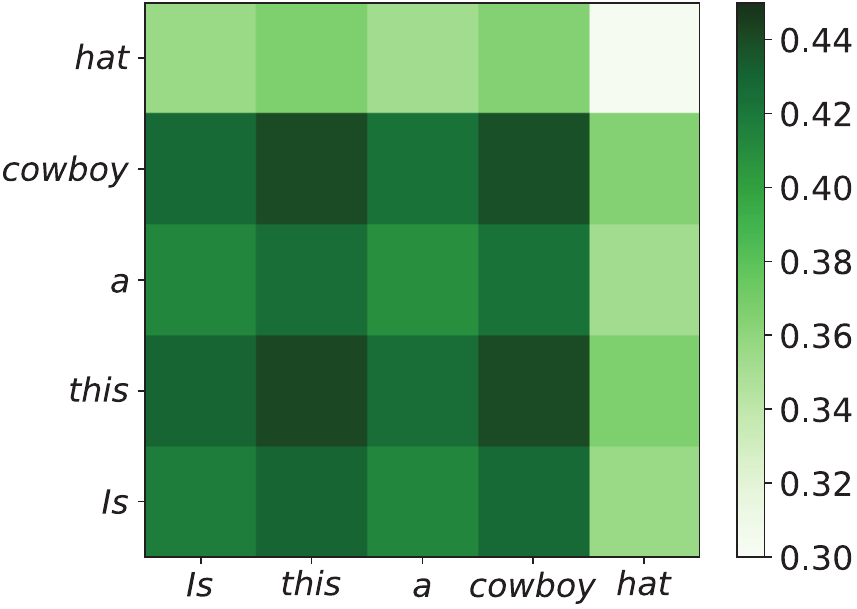}}
        \label{fig:short-a}
    	% \hfill
        % \hspace{1.cm} 
    	\subfloat[]{\includegraphics[width = 0.25\textwidth]{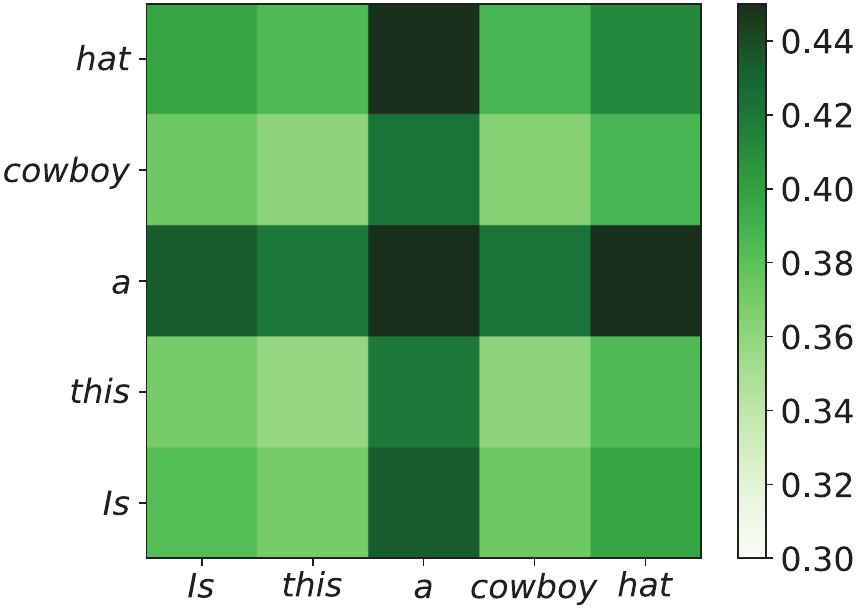}}
        \label{fig:short-b}
        % \hspace{1.cm} 
    	\subfloat[]{\includegraphics[width = 0.24\textwidth]{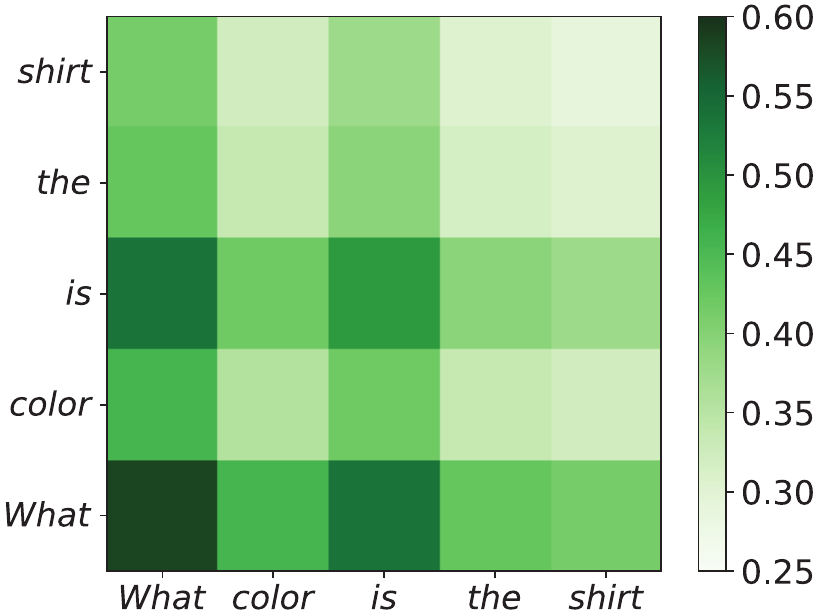}}
        \label{fig:short-c}
        % \hspace{1.cm} 
    	\subfloat[]{\includegraphics[width = 0.24\textwidth]{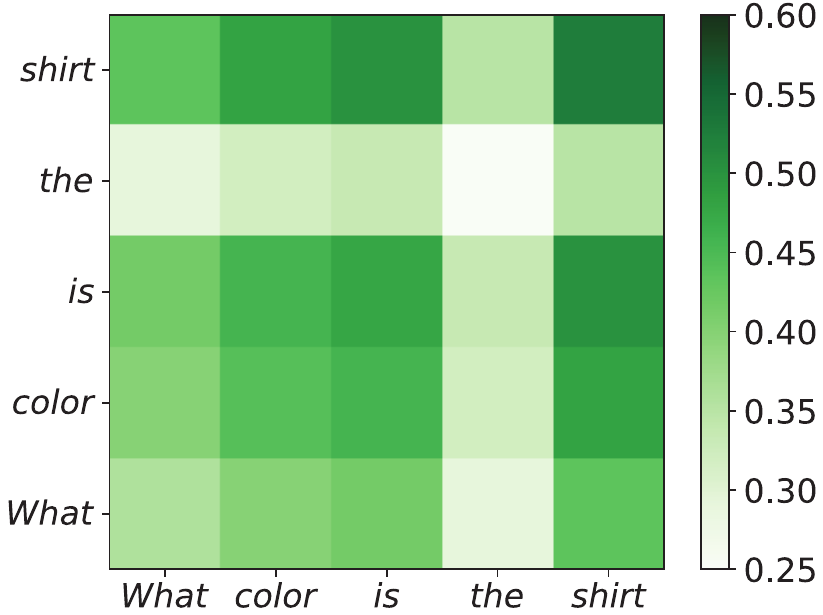}}
        \label{fig:short-d}
    \caption{Examples of the visualization concerning the weight mapping to each word of the questions. The subplots (a) and (b) are the visualizations with respect to the question \textit{``Is this a cowboy hat?''}, (a) is the result of the UpDn model trained with original questions, (b) is the result of UpDn model trained with variant questions. (c) and (d) are the visualizations with respect to the question \textit{``What color is the shirt?''}, (c) is the result of UpDn model trained with original questions, (d) is the result of UpDn model trained with variant questions.}
    \label{Fig:3}
    \end{figure*}
    \begin{figure}[!h]
    	\centering
    	\includegraphics[width =.48\textwidth,height=0.25\textwidth]{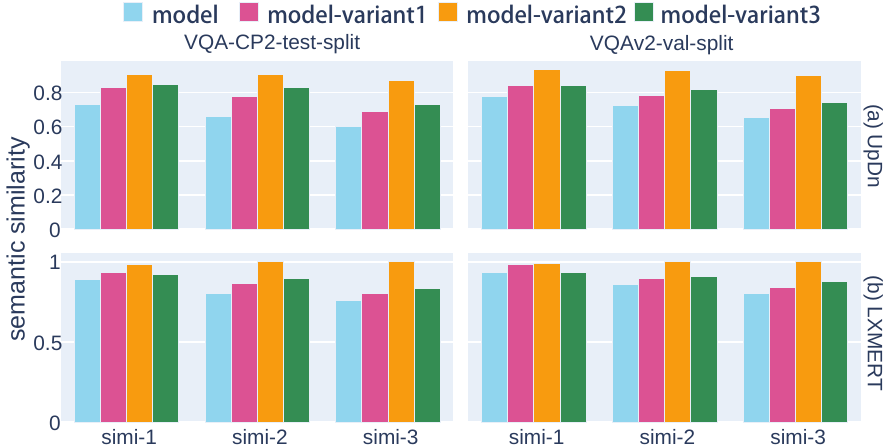}
    	\caption{The semantic similarity between the encoded original questions and the encoded variant questions under different models. simi-(i) means the semantic similarity between the original question and the variant-(i) question. }  
    	\label{Fig: simi}
     % \vspace{-0.5cm}
    \end{figure}
    In addition, more detailed results are presented in Table 3. For instance, the results of the trained variant models on the in-distribution VQAv2 dataset reveal that almost every model's performance, in terms of each answer type, has decreased to varying degrees when compared to the performance of the original models. This demonstrates that learning the pattern of the variant questions would negatively affect the performance of the original question on in-distribution data.
    However, the performance of variant models on the OOD test set is quite different. Almost all models showed improvements in the All metric, primarily due to the improvements in ``Yes/No'' and ``Num'' answer types. However, the models trained with variant-1 questions, including Q-only, SAN, BAN, and UpDn, showed a decrease in performance on the ``Other'' metric. In contrast, LXMERT improved its performance when trained with variant-1 questions, with an accuracy improvement of +1.88\% on ``Other'' metric. The reason for these experimental results may lie in the different question encoders used by these models.

\subsection{Other Property}
    Furthermore, we found that the trained variant models exhibit better semantic robustness than the original models.
    We calculated the semantic similarity between the encoded original questions and encoded the variant questions by:
    \begin{equation}
    simi = 1- \frac{1}{N} \sum_{i}^{N} cos<q_i, var_i>
    \end{equation} 
    where $N$ is the number of the samples, $q_i$ and $var_i$ denote the encoded original question and variant question, respectively. The results are shown in Figure 4.

\section{How to Utilize These Traits?}
The variant models exhibit more promising results on VQA-CPv2 compared to the original model. As demonstrated in the previous section, the variant models can avoid learning the inherent prior knowledge related to question types from the questions, allowing them to focus on other useful patterns. However, it should be noted that the syntactic structure of the variant questions may be incomplete or incorrect, and the semantics of the variant questions may even be entirely different. Therefore, the variant questions can only be utilized to aid in the design of de-biasing methods.

\subsection{Proposals}
    To take advantage of the trait of variant models while preserving the semantics of the original question, one approach is to use the contrastive learning paradigm to combine the two encodings. In this approach, the variant question is treated as the positive sample, while negative samples are randomly sampled from the mini-batch during training. The process can be formulated as $\mathcal{L}_{con} = -\frac{1}{N} \sum_{i}^{N}log(\frac{e^{sim(h_i,h^{pos}_i)}}   {e^{sim(h_i,h_i^{pos})}+ \sum{e^{sim(h_i,h_i^{neg})}}})$,
    where $h_i$ is the joint feature representation of two kinds modality, $h^{pos}_i$ and $h^{neg}_i$ represent the positive feature and negative feature, respectively, 
    $sim(h_i,h^{pos}_i)=\frac{h_i\cdot h^{pos}_i}{||h^{pos}_i|| \cdot ||h_i||}$.
    During the training stage, the total optimization objective includes
    the VQA loss $\mathcal {L}_{ce}$
    and the contrastive loss $\mathcal{L}_{con}$.
    Besides, another straightforward way is to combine the features of the original question with those of the variant question directly as a kind of data augmentation. 
    Specifically, the joint features were encoded by a weighted combination of two kinds of features. Therefore, the resulting features cover richer patterns.

\subsection{Experiments}
    In this section, we validate the effectiveness of the latter de-biasing method on the OOD benchmark, \textit{i.e.,} VQA-CPv2.
    The proposed method is model-agnostic and can be combined with any other VQA model. Here, we also choose the most widely used base VQA models, SAN, UpDn, BAN, LXMERT, MCAN, and the Q-only model, as the baseline models. Regarding the implementation details in the training process, we adhere to the experimental settings of the open-source codes and do not modify other parameters such as learning rate, batch size, or optimizer.
    The experimental results are presented in Table 5. As evident from the results, all base models exhibited performance improvements when integrated with the method proposed in this paper. Moreover, the majority of models demonstrated enhancements in the ``Other" metric, with only the UpDn and LXMERT models experiencing a slight decrease.
    This indicates that the base models combined with the proposed method can not only improve the simple pattern but also learn more difficult patterns.
    In addition, while the performance of some models combined with the proposed method may not be as good as the results of the variant models, the overall accuracy has improved significantly compared to the original models. 
    \begin{table}[!ht]
          \centering
          \setlength{\tabcolsep}{7.5pt}
          \renewcommand{\arraystretch}{1.05} 
            \begin{tabular}{l|cccc}
            \hline
            \multirow{2}{*}{Model} & \multicolumn{4}{c}{VQA-CPv2} \\
        \cline{2-5}          & All   & Yes/No   & Num   & Other \\
            \hline
            Q-only & 21.37 & 41.01 & 12.14 & 13.61 \\
            Q-only+ours & {\textbf{26.3}} & 42.59 & 11.7  & 16.23 \\
            \hline
            SAN   & 40.70  & 41.62 & 13.14 & 47.77 \\
            SAN+ours & {\textbf{41.41}} & 43.37 & 12.82 & 48.23 \\
            \hline
            UpDn & 41.53 & 42.91 & 13.56 & 48.55 \\
            UpDn+ours & {\textbf{44.95}} & 54.51 & 14.89 & 48.18 \\
            \hline
            BAN   & 41.73 & 42.72 & 13.51 & 48.95 \\
            BAN+ours & {\textbf{43.63}} & 45.96 & 15.04 & 50.25 \\
            \hline
            MCAN  & 43.73 & 42.6  & 15.69 & 52.02 \\
            MCAN+ours & {\textbf{44.89}} & 44.58 & 16.24 & 52.92 \\
            \hline
            LXMERT  & 43.29 & 46.37  & 15.38 & 49.43 \\
            LXMERT+ours & {\textbf{50.85}} & 71.54 & 17.19 & 49.24 \\
            \hline
            \end{tabular}%
            \caption{The experimental results (Acc.\%) of base VQA models combined with the proposed data augmentation method. Note that the LXMERT was fine-tuned for 20 epochs.}
          \label{tab5}%
           \vspace{-0.4cm}
    \end{table}%

\section{Conclusion and Future Work}
    In this paper, we investigate language modality in the VQA task through experimental analysis. The empirical findings indicated that the issue of language priors bias is not only related to question types alone, the postfix of questions even has a greater impact on language bias. Furthermore, we observed that variant models outperform original models on the VQA-CPv2 benchmark. We identified the underlying reasons for these results and proposed new debiasing methods based on these findings. The experimental results demonstrated that our method enhances the VQA models' generalization ability. Our main purpose is not to pursue state-of-the-art results but to gain insights for designing bias-reduction methods. However, we only present some novel experimental findings, and we plan to provide in-depth theoretical analysis and probe other methods to leverage these traits in future work.

\section*{Acknowledgments}
    This work was supported in part by the National Natural Science Foundation of China under Grant No. 62276110, No. 62172039 and in part by the fund of Joint Laboratory of HUST and Pingan Property \& Casualty Research (HPL). The authors would also like to thank the anonymous reviewers for their comments on improving the quality of this paper.

\bibliographystyle{named}
\bibliography{ijcai23}
\end{document}